\pdfoutput=1

\documentclass[11pt, table]{article}
\usepackage{pdfpages}

\usepackage{colortbl}
\usepackage{adjustbox}

\usepackage{rotating}
\usepackage{tablefootnote}
\usepackage{coling}
\usepackage{tabularx}
\usepackage{arydshln}
\usepackage{listings}
\usepackage{booktabs}
\usepackage{multirow}
\usepackage{multicol}
\usepackage{times}
\usepackage{latexsym}
\usepackage{makecell, xltabular}
\usepackage[most]{tcolorbox}
\usepackage{hyperref}
\usepackage{url}
\usepackage{paralist}
\usepackage{subcaption}
\usepackage{lscape}
\usepackage{array}
\usepackage{graphicx}
\usepackage{tabularx}
\usepackage{paralist}
\usepackage{tikz,lipsum,lmodern}
\usepackage[most]{tcolorbox}

\usepackage[skip=0.44\baselineskip]{caption}
\usepackage[T1]{fontenc}

\usepackage[utf8]{inputenc}

\usepackage{microtype}

\usepackage{inconsolata}

\usepackage{geometry}
\usepackage{array}
\usepackage{amssymb}
\usepackage{adjustbox}
\usepackage{lscape}
\usepackage{hyperref}
\usepackage[utf8]{inputenc}

\title{On the effective transfer of knowledge from \\ English to Hindi Wikipedia}

\author{
  Paramita Das \\
  IIT Kharagpur \\
  \texttt{dasparamita1708@gmail.com} 
  \And
  Amartya Roy \\
  Bosch \\
  \texttt{Amartya.Roy@in.bosch.com} 
  \AND 
  Ritabrata Chakraborty \\
  Osmania University \\
  \texttt{ritabratac1997@gmail.com}
  \And
  Animesh Mukherjee \\
  IIT Kharagpur \\
  \texttt{animeshm@cse.iitkgp.ac.in} 
}

\begin{document}
\maketitle

\begin{abstract}
Although Wikipedia is the largest multilingual encyclopedia, it remains inherently incomplete. There is a significant disparity in the quality of content between high-resource languages (HRLs, e.g., English) and low-resource languages (LRLs, e.g., Hindi), with many LRL articles lacking adequate information. To bridge these content gaps we propose a lightweight framework to enhance knowledge equity between English and Hindi. In case the English Wikipedia page is not up-to-date, our framework extracts relevant information from external resources readily available (such as English books), and adapts it to align with Wikipedia's distinctive style, including its \textit{neutral point of view} (NPOV) policy, using in-context learning capabilities of large language models. The adapted content is then machine-translated into Hindi for integration into the corresponding Wikipedia articles. On the other hand, if the English version is comprehensive and up-to-date the framework directly transfers knowledge from English to Hindi. Our framework effectively generates new content for Hindi Wikipedia sections, enhancing Hindi Wikipedia articles respectively by 65\% and 62\% according to automatic and human judgment-based evaluations.
\end{abstract}

Despite the wide usage of the multilingual content of Wikipedia, significant knowledge gaps exist across different language editions of Wikipedia~\cite{miquel2018wikipedia}, creating an information divide. For instance, the Hindi Wikipedia, with only 163 thousand articles as of 2023, contrasts sharply with the English Wikipedia's 6.8 million articles\footnote{\url{https://en.wikipedia.org/wiki/List_of_Wikipedias}}, despite Hindi being the third most spoken language globally. It is evident that in many low-resource languages (LRLs), Wikipedia often lacks pages on important global topics or notable individuals due to the limited participation of community editors. This disparity limits the engagement of LRL communities with online resources and educational content. Moreover, Wikipedia articles on the same topic often differ significantly across languages~\cite{miquel2020wikipedia,roy2020topic} due to factors such as cultural relevance and the varied expertise of contributors. Addressing these disparities is crucial for achieving \textit{knowledge equity}, a concept emphasized by the Wikimedia Foundation~\cite{redi2020taxonomy}. Existing studies~\cite{zhang2024retrieval,shao2024assisting} primarily focus on generating full-length Wikipedia articles in English, which restricts research efforts for LRLs. Recent studies~\cite{taunk2023xwikigen,shao2024assisting,maurya2023towards} have made significant progress in automated methods for cross-lingual knowledge transfer, particularly in generating full-length articles in LRLs. However, many of the existing approaches focus on generating articles from scratch, which is often less effective for enriching existing articles and overlooks the collaborative nature of knowledge creation. On large crowd-sourced platforms, such as Wikipedia, collaborative efforts, especially human-curated content hold greater importance than automatically generated information. Often, Wikipedia articles on specific topics in LRLs require enrichment in certain sections compared to their counterparts in high-resource languages (HRLs). To the best of our knowledge, no prior work has specifically addressed the enrichment of section-specific content of Wikipedia articles in LRLs. To address this issue, we propose a lightweight approach that enriches section-specific contain while preserving existing human-authored material and augmenting it with carefully integrated, automatically generated knowledge using standard NLP techniques. Our framework utilizes the standard retrieval augmented generation (RAG) framework to extract relevant information from a web corpus for a given section title, followed by machine translation of the extracted content into LRL. Additionally, the framework addresses additional challenges, especially finding suitable references relevant to the section. In our experiment, we consider English as a representative HRL and Hindi as a representative LRL. The key research question drives our experiment: How can we automatically transfer knowledge from the more enriched language version (HRL) to the less enriched one (LRL) in Wikipedia, given a Wikipedia article on a topic $t$? Our contributions are as follows.
\begin{compactitem}
    \item We propose a multistage approach to extract knowledge from biographical writings about popular figures, transform this text to adhere to the NPOV guidelines and incorporate it into low-resource Wikipedia articles, e.g., Hindi. This approach uses the \textit{WikiTransfer} framework to identify and transfer relevant content from English Wikipedia to Hindi Wikipedia, leveraging IndicTrans\footnote{\url{https://ai4bharat.iitm.ac.in/indic-trans2/}}.
    \item We manually curate 103 biographical writings relevant to corresponding Wikipedia articles as external knowledge sources, which can serve as a rich source of factual information.
    \item Our lightweight framework efficiently updates Hindi Wikipedia articles by adding coherent and new information. Our rigorous evaluation through both automated and crowd-based assessments demonstrates an improvement of 65\% and 62\%, respectively, in terms of integrating new information. Our code and dataset for reproducing similar content are available at \url{https://github.com/paramita08/wikiTransfer}.
\end{compactitem}

\noindent The applicability of our framework has been demonstrated within the Wikipedia domain; however, its potential usage extends to large-scale industrial applications where enriching local knowledge repositories using open-source automated systems, such as large language models (LLMs), is infeasible due to proprietary data or domain-specific constraints. In such scenarios, our pipeline—leveraging semantic search based retrieval systems, i.e., RAG, and further adapting the system to be used in low-resource settings through domain-sensitive LLMs \cite{liu2023chipnemo}—offers a practical and scalable solution. Our approach is particularly relevant for industries dealing with sensitive or specialized knowledge repositories, where conventional generative AI models may fall short.

\section{Related work}
\label{sec:related_work}
The research community has expanded NLP research in multilingual settings including more languages, especially non-English, and smaller low-resource language editions~\cite{wang2023all}. In case of Wikipedia, many researchers have examined differences between different language editions of Wikipedia from the standpoints of content (i.e., text, image), readers~\cite{arora2022wikipedia}, and editors~\cite{bipat2018we} as well. Text diversity in Wikipedia has collectively demonstrated that textual content about the same topic is highly diverse across language editions~\cite{hecht2010tower,roy2020topic}. Different language editions of Wikipedia serve very different communities~\cite{johnson2021global,lemmerich2019world} and thus often cover very different topics~\cite{paramita2017using}. This information gap results in variations in the quality and quantity of content~\cite{lewoniewski2017relative}, presumably affecting the vocabulary and ability of language models trained on Wikipedia to handle different topics accurately. A large body of work~\cite{adar2009information, wulczyn2016growing, paramita2017using} based on vanilla NLP approaches tried addressing the information asymmetry between different language editions. With the advancement of generative AI, recent works~\cite{agarwal2020wikipedia,shivansh2023cross,guo2024teaching} have focused on content alignment and content transfer in low-resource languages from scratch. In the case of languages with limited or poor translation resources, authors in~\cite{paramita2017using} proposed a lightweight approach to measure cross-lingual similarity in Wikipedia using section headings rather than the entire Wikipedia article, and language resources derived from Wikipedia and Wiktionary to perform translation. Although existing research works are valuable, but lacks an end-to-end pipeline for transferring content from high-resource to low-resource languages, limiting efforts to bridge Wikipedia's content gap. Our work addresses this issue.

\section{Dataset description}\label{sec:dataset}
We employ a systematic approach to collect Wikipedia articles, which are available in both English and Hindi versions. We also anchor on the content of external resources for a subset of articles. In this work, we utilize a dataset comprising biographies, i.e., articles of Wikipedia category \textit{people} sourced from Wikipedia. As the biography articles in Wikipedia follow a predefined structure across multiple languages, we concentrate on biography articles of renowned persons as the dataset for our experiment. Although our experiments have focused solely on biography articles, our framework is versatile and can be readily applied to other types of Wikipedia articles, such as articles covering technical concepts or geographical locations. This extension is feasible as long as a sufficient digital corpus on the topic is available to serve as an external resource for the RAG module of our pipeline.\\
\noindent\textbf{Collection of Wikipedia articles:} Authors in~\cite{Beytía_Agarwal_Redi_Singh_2022} published a dataset of Wikipedia biographies in the ten most widely spoken languages, including English and Hindi\footnote{\url{https://www.visualcapitalist.com/100-most-spoken-languages/}}. We use this dataset to extract biographies, along with their Wikidata IDs, which serve as unique identifiers across language versions. For instance, Serena Williams' biography has the Wikidata ID \textit{Q11459}, allowing retrieval in all available languages. First, we collected a set of 21,340 biography articles in both Hindi and English versions from the above-mentioned dataset. Using the MediaWiki API\footnote{\url{https://www.mediawiki.org/wiki/API:Get_the_contents_of_a_page}}, we retrieve and pre-process the current version of wikitext of these articles. Section headings are extracted using the Wikipedia Python package\footnote{\url{https://pypi.org/project/wikipedia/}}, excluding sections like \textit{See also}, \textit{References}, and \textit{External links}, and \textit{Inline citations}.\\
\noindent\textbf{Collection of article quality:} We utilize article quality as an indicator to determine which language version (English and Hindi) contains more enriched information between the two. Therefore, we aim to gather the quality scores for each language version of Wikipedia articles. Using the dataset from~\cite{das2024language}, we collect quality scores (ranging from 0 to 1) for Hindi and English Wikipedia articles. We identify a subset of 17,226 articles where Hindi scores are lower than English, serving as our candidate set for further experiments. Since no rigid quality class hierarchy exists for Hindi articles on Wikipedia, we use this language-agnostic quality score. Next, we extract quality categories (FA, A, GA, B, C, Start, Stub) for English articles using \textsc{ORES}. For articles in the FA category (according to English Wikipedia), we directly use their content to improve lower-quality Hindi versions. For other categories, English articles are first enhanced using external resources, followed by transferring the improved content to Hindi, thus ensuring that the highest-quality information is used to enhance Hindi articles. \\
\noindent\textbf{Collection of external resources:} We use online digital library \textit{Archive}\footnote{\url{www.archive.org}}, to source biographical writings for our enhancements. \textit{Archive} offers a vast collection of scanned historical books, making it ideal for our needs.\\
\noindent\textit{Automated search:} We first construct a search query using the title of each biography article to locate the corresponding page on \textit{Archive}. For a given biography, say $P$, the query retrieves biographical writings, say \texttt{bio}. We use the requests library and the HTTP GET method to extract the page content. If the keyword 'biography' is found, the response is considered valid, ensuring relevant results from \textit{Archive}.\\
\noindent\textit{Manual verification}: Due to name ambiguity and automated search limitations, many results contain irrelevant information. To address this, a post-graduate student who frequently uses Wikipedia manually verified the collected links. This ensures the quality and relevance of the biographical writings used. We download the verified biographical writings in text format to enrich Wikipedia articles, aiming to improve the quality of both English and Hindi biographies. Thus, our curated dataset includes Wikipedia articles in both English and Hindi, their quality scores, and a set of biographical writings extracted from external sources. The dataset statistics is noted in Table~\ref{tab:dataset_stat}.
\begin{table}[t]
\scriptsize
    \centering
    \begin{tabular}{|c|c|c|}
    \hline
    \rowcolor{green!20}
       Quality class & \# of Articles & \# of Biographies \\
       \hline
       FA  & 235 & 0  \\
       \hline
       A & 6 & 0 \\
       \hline
       GA & 485 & 13  \\
       \hline
       B & 1930 & 51 \\
       \hline
       C & 3428 & 38 \\
       \hline
       Start & 6625 & 0 \\
       \hline
       Stub & 4517 & 0 \\
       \hline
    \end{tabular}
    \caption{\footnotesize Filtered dataset -- articles categorized in quality classes and biographical writings extracted for the corresponding classes.}
    \label{tab:dataset_stat}
\end{table}

\section{Experimental pipeline}
\label{sec:method}

We propose an end-to-end pipeline to transfer knowledge from English articles to their corresponding Hindi versions. The pipeline includes the framework WikiTransfer and additional modules for external knowledge augmentation and POV correction to enhance the Hindi version of articles.

\subsection{WikiTransfer}
WikiTransfer first identifies semantically similar section titles between English and Hindi Wikipedia articles. To map sections, we translate Hindi titles to English using IndicTrans, compute embeddings for all titles, and measure cosine similarity between pairs of Hindi and English titles of every article. For instance, for a Hindi title denoted as $t_{h}$, we compute the similarity with the embedding of $m$ English section titles of an article $p$. For each Hindi title, the most similar English section is identified as the source for content transfer. We use the sentence transformers model all-MiniLM-L12-v2\footnote{\url{https://huggingface.co/sentence-transformers/all-MiniLM-L12-v2}} for embedding computation. Section pairs with a similarity score above a threshold of 0.44 (mean similarity) are selected as mapped sections. After matching section titles, we analyze the content of the mapped sections for coherence. We compute embeddings of the section content (Hindi and English) using multilingual e5-large\footnote{\url{https://huggingface.co/intfloat/multilingual-e5-large}} and calculate cosine similarity. Section-content pairs with similarity scores above a threshold of $\mu + \sigma$ (mean: 0.89, std dev: 0.06) are considered similar. \\ 
\noindent\textbf{Content augmentation:} After mapping sections and content between English and Hindi Wikipedia articles, this step involves translating English content into Hindi using the IndicTrans model~\cite{gala2023indictrans2}. Translated sentences are appended to the existing content in the mapped Hindi section. Before incorporating the translated sentences, we apply a two-step filtering: (1) discard short translations with one or two words to avoid errors, and (2) use the multilingual e5-base model of the sentence transformer to identify the top three semantically related sentences for each existing Hindi sentence. Likewise previous mapping scheme, a cosine similarity score is calculated for each $x$ in the existing Hindi sentences with individual translated sentences in $Hindi(e)$, and we select the top three sentences among the sorted (in descending order) $Hindi(e)$ sentences that belong to the range of $\mu$ and $\mu + \sigma$ of the similarity scores. This way, we pick up three sentences that are somewhat dissimilar from the existing sentence $x$, thus avoiding redundancy. If a sentence is selected for multiple matches, it is appended only once, creating a reduced and informative set of translated sentences to be added.

\subsection{Content extraction from biography}
English articles that belong to quality classes other than FA might require additional information to enhance their quality. Therefore, we first gather information from external resources of biography writings (as described in section~\ref{sec:dataset}) for such articles in our dataset and add them to the appropriate sections for further processing using the WikiTransfer framework. To extract information from these external narratives aligned with the content of the articles, we employ the standard RAG method. For each English article, we select the most recent biography, split the text into chunks using the \textit{RecursiveTextSplitter}\footnote{\url{https://github.com/langchain-ai/langchain}} function, and embed each chunk using \texttt{sentence-bert}\footnote{https://huggingface.co/sentence-transformers/all-mpnet-base-v2} embeddings which are stored in the vector database--\textsc{ChromaDB}. Now, for the given English article $E_{p}$ with $m$ sections, we provide the content of the section $t_{i}$ where $i \in {1,2,..,m}$ as query and external narratives $W$ as the input to the \textit{retriever} module of the RAG pipeline. We use maximum marginal relevance (MMR) based search to retrieve top $k$ chunks (we fix $k$ to 3) relevant to the query. Out of these retrieved chunks, we utilize a suitable prompt (Llama-3(8B)-Instruct model), which identifies which chunk is the most relevant to the given section content (please see the prompt in Appendix [A]). Instead of using the RAG text generator, we use a POV rectifier module (as discussed below) to refine the content.

\subsection{POV correction}
Alongside Wikipedia's openness, a fundamental pillar of its success is its commitment to the NPOV policy, which ensures that facts should be presented fairly and impartially. According to this policy, Wikipedia prohibits sentences that contain perspective-specific or biased language, such as expressions of praise, criticism, or other sentiments that reflect the editor's personal feelings. Given that we are extracting content from biographies, which may include subjective language, there is a risk that the new content could violate Wikipedia's NPOV\footnote{\url{https://tinyurl.com/cb7yv3tt}} standards. Therefore, to adhere to Wikipedia's NPOV policy, we identify and remove subjective biases, named as framing bias and epistemological bias~\cite{recasens2013linguistic} from individual sentences extracted from the biographies if they exist and rephrase them accordingly. In this study we have tried to leverage the power of LLMs for the generation of Wiki-style content. The most popular methods to use LLM in such downstream tasks are $\bullet$ supervised finetuning \cite{jiang2019smart} $\bullet$ in-context learning \cite{sahoo2024systematic}. We have performed our experiments with LLama-3(8B) instruct model \cite{llama3modelcard} for both these setups.\\
\noindent\textbf{Supervised fine-tuning (SFT)}: For this setup, we have fine-tuned the LLama-3(8B)-instruct model using the WNC and WikiBias \footnote{We have used $\sim2k$ from WikiBias ($Train_{manual}$) \cite{zhong2021wikibias} and 10k biased and neutral sentence pairs randomly sampled from WNC corpus~\cite{pryzant2020automatically} as training data.} corpus in a supervised fashion to obtain a supervised fine-tuned model.\\
\noindent\textbf{In-context learning (ICL)}: For this setup, we have used off-the-shelf instruction-tuned models namely LLama-3(8B) \& LLama-3(70B). We have used a generic prompt (please see the prompt in Appendix [B]) to generate a debiased sentence given a biased sentence. Specifically, we have tried $\bullet$ zero-shot (only the instruction) and $\bullet$ few-shot~\cite{parnami2022learning} (a few examples are used to describe the task to the model) prompting for the generation of Wiki-style neutral content.\\
\noindent\textbf{Evaluation}: Both of these configurations are conducted on a sample of test data comprising 431 biased sentences and their neutral counterparts\footnote {We have utilized WikiBias test data here.} and the evaluation of generated neutral content are computed using three reference-based metrics--
BLEU~\cite{papineni2002bleu}, METEOR~\cite{banerjee2005meteor},
and BERTScore ~\cite{zhang2019bertscore}. As evidenced in Table \ref{tab:model_benchmark}, ICL consistently outperformed SFT across all three the reference-based metrics. Hence we have used the ICL few-shot setup (5 examples have been used in the prompt) and Llama-3(70B) in generating neutral content for the extracted external book content as mentioned in the above section.
\begin{table}[h]
\scriptsize
\centering
\renewcommand{\arraystretch}{1.5} 
\begin{tabular}{|p{2.2cm}|p{0.9cm}|p{0.7cm}|p{1.1cm}|p{0.7cm}|}
\hline
\rowcolor{green!20} 
Model & Methods & BLEU & METEOR & BERT \\ \hline
\multirow{2}{*}{Llama-3(8B)[SFT]} & Zero-shot & 0.27 & 0.5 & 0.94 \\ \cline{2-5} 
                                 & Few-shot  & 0.35 & 0.65 & 0.92 \\ \hline
\multirow{2}{*}{Llama-3(8B)[ICL]} & Zero-shot & 0.25 & 0.6 & 0.93 \\ \cline{2-5} 
                                 & Few-shot  & 0.35 & 0.66 & 0.95 \\ \hline
\multirow{2}{*}{Llama-3(70B)[ICL]}& Zero-shot & 0.24 & 0.57 & 0.93 \\ \cline{2-5} 
                                 & Few-shot  & \fbox{0.4} & \fbox{0.68} & \fbox{0.95} \\ \hline
\end{tabular}
\caption{\footnotesize Evaluation score of Llama-3 on test data.}
\label{tab:model_benchmark}
\end{table}

\begin{table*}[t]
\tiny
\centering
\begin{tabular}{|p{0.5cm}|p{7.8cm}|p{6.8cm}|}
\hline
\multicolumn{3}{|c|}{\textbf{Original biased sentence}: Blacks never listen to their parents.} \\ \hline
\textbf{Score} & \textbf{Rules}                                                           & \textbf{Example}                                                                       \\ \hline
\textbf{\textcolor{red}{1}}           & Complete bias removal & People do not always listen to their parents. \\ \hline
\textbf{\textcolor{blue}{2}}          & Complete bias removal + Keeping the meaning (context) same & Some people never listen to their parents. \\ \hline
\textbf{\textcolor{green}{3}}         & Complete bias removal + keeping the meaning (context) same + fluency & It is not uncommon for individuals to disregard parental advice. \\ \hline
\end{tabular}
\caption{\footnotesize\textbf{Scoring metric.} Details of the scoring metrics used for annotation along with examples based on a biased sentence. The original biased sentence is taken from CrowS-Pairs dataset~\cite{nangia2020crows}.}
\label{tab:scoring_metric}
\end{table*}
\begin{table*}
    \scriptsize
    \begin{tabular}{|c|c|c|c|c|c|c|c|c|}
    \hline
    \rowcolor{green!20}
    \textbf{Type} & \multicolumn{2}{c|}{\textbf{Info}} & \multicolumn{2}{c|}{\textbf{Read}} & \multicolumn{2}{c|}{\textbf{Und}}  & \multicolumn{2}{c|}{\textbf{Qual}} \\ \hline
    & $c_{old}$ & $c_{new}$ & $c_{old}$ & $c_{new}$ & $c_{old}$ & $c_{new}$ & $c_{old}$ & $c_{new}$\\ \hline
    FA & 53.35(50.65) & 114.02(73.84) & 4.66(0.79) & 4.90(0.62) & 17.09(4.02) & 18.05(2.84) & 26.17(13.69) & 42.39(19.04) \\ \hline
    Non-FA & 62.94(45.09) & 136.17(61.7) & 4.92(0.72) & 5.17(0.62) & 17.01(3.65) & 17.74(2.75) & 28.74(11.76) & 47.99(15.66)\\ \hline
   \end{tabular}
    \caption{\footnotesize Human evaluation on the generated machine-translated Hindi content based on three metrics.}
    \label{tab:auto_eval}
\end{table*}

\section{Results}\label{sec:result}

We assess the LLM-generated neutral Wiki-style content and the machine-translated Hindi content through automatic metrics and human evaluation. 

\subsection{Automatic evaluation}
To evaluate the relatedness and quality of the newly generated content with the pre-existing Hindi text, we employ the E-A-T framework proposed by~\cite{sugandhika2022assessing} for the information quality assessment of Wikipedia content, and three important factors of the framework are -- Expertise ($E$), Authority ($A$), and Trust ($T$). For the purpose of assessment of the machine-translated content, we valued $E$ more than the other two factors, which comprise the following components --\\
\footnotesize
\underline{Informativeness} ($Info$) = 0.12 * \texttt{page-size} + 0.151*\texttt{\#sentences} + 0.154 * \texttt{\#words} + 0.155 * \texttt{\#complex-words};\\
\noindent \underline{Readability} ($Read$) = 0.213 * \texttt{Flesch-Kincaid-grade-evel} + 0.185 * \texttt{Coleman-Liau-index} + 0.26 * \texttt{\%complex-words} + 0.253 * \texttt{avg-syllables-per-word};\\ 
\noindent \underline{Understandability} ($Und$) = 0.393 * \texttt{Gunning-Fog-score} + 0.352 * \texttt{SMOG-index} + 0.181 * \texttt{automated-readability-ndex} + 0.344 * \texttt{avg-words-per-sentence}; \\
\normalsize
\noindent Finally, $E$ is measured in terms of the Quality of a Wikipedia page content which is defined as: \underline{Quality} ($Qual$) = 0.255 * Informativeness + 0.654 * Readability +  0.557 * Understandability. Informativeness represents the size of the textual content on the Wikipedia page; readability and understandability provide insights into the linguistic quality of the article content.  We perform reverse translation of newly generated and existing Hindi content into English and then evaluate it using the above-mentioned approach.\\ 
\noindent \textbf{Results of automatic evaluation}: Due to limited resources for evaluating Hindi text quality, we assess the quality in English. We perform reverse translation of both the newly generated Hindi content $c_{new}$ and the existing Hindi content $c_{old}$ (which is the set of existing Hindi sentences in a section) and then evaluate them using the above-mentioned approach. We compute the metrics for individual sections and average over all the sections of the articles under consideration. The scores obtained for the two groups -- (1) FA and (2) non-FA (GA, B, and C quality articles together) using automatic evaluation are tabulated in Table~\ref{tab:auto_eval}. Overall, we observe that the enhanced content is superior to the old content in terms of all the metrics for both groups. Since the standard deviation obtained in the case of Informativeness is large, we provide further division of the metrics (mean and standard deviation) in Table~\ref{tab:auto_eval_FA_ranges} in the Appendix.  \\

\subsection{Human evaluation}
\noindent\textbf{Assessment of LLM-generated NPOV text}: To evaluate the neutrality of the LLM-generated text, we conduct the human assessment on 50 randomly sampled sentences from our dataset, comparing the original sentence from external resources with the NPOV version generated by the Llama-3(70B) model. Two evaluators assigned scores using the scoring metric defined in Table \ref{tab:scoring_metric}. The average score ($Score_{neu}$) assessed by the two evaluators are 82.85\% and 77.14\% , respectively, showing that the neutralized content is suitable for augmenting the target Hindi sections. \\
\noindent\textbf{Overall assessment}: We evaluate the content generated by our pipeline in two scenarios: (a) augmenting content using only \textit{FA} articles, and (b) augmenting content from non-\textit{FA} articles along with external sources. The evaluation focuses on three qualitative metrics -- \textit{informativeness}, \textit{readability}, and \textit{coherence} -- each rated on a scale from 1 to 3. Informativeness, in this context, indicates the ability of a piece of text to provide useful information and comprehensive content. Readability measures the effort required by the reader to read and understand a piece of information. If the vocabulary and sentence structure in the text are complex, the difficulty of reading increases. Coherence represents the logical flow between sentences in a text, ensuring that they naturally follow one another to form meaningful content. \textit{Seven} Hindi-speaking evaluators conducted this assessment. For each metric, they compared the original Hindi content, $c_{old}$, consisting of existing Hindi sentences ($h$), with the newly generated Hindi content, $c_{new}$, of the corresponding section. Evaluators assigned scores based on improvement, no change, or decline, labeled as 3, 2, and 1, respectively where a higher score indicates greater improvement of $c_{new}$ compared to $c_{old}$. We randomly sample 35 sections from our dataset to evaluate the content generated by our framework: 10 from the FA category and 25 from non-FA quality categories (5 GA, 10 B, and 10 C). The average informativeness, readability and coherence across the seven human judges respectively are -- (FA: 2.67, non-FA: 2.68), (FA: 2.46, non-FA: 2.50), and  (FA: 2.34, non-FA: 2.32)\footnote{Please see Table~\ref{tab:human_eval} in the Appendix for ratings obtained for each of the seven individual annotators.}. Thus we observe that for all the metrics and for both FA and non-FA categories, the average judgements are always 2.3+ indicating the newly generated content is reasonably good in terms of the three metrics, especially informativeness and readability. We find significant improvement in informativeness for both FA and non-FA groups, suggesting effective addition of relevant knowledge to existing sections (see Appendix [Figure~\ref{fig:sample_FA_hindi}, Figure~\ref{fig:sample_FA_english}, Figure~\ref{fig:sample_C_hindi}, Figure~\ref{fig:sample_C_english}] for examples). Given our multi-label evaluation scheme and involving multiple annotators, we compute inter-annotator agreement using Fleiss' Kappa method~\cite{fleiss1973equivalence}. We obtain $\kappa$ values of 0.61, 0.53, and 0.54 for informativeness, readability, and coherence, respectively, indicating a moderate to substantial level of agreement among annotators in assessing the generated content.

\section{Conclusion} \label{sec:conclusion}
In this paper, we presented a lightweight framework to produce content that is substantially superior to the existing content for Hindi articles. From FA-quality English articles, we directly translated relevant content to their corresponding Hindi counterparts. For non-FA articles, we first extracted relevant content from external sources and adapted these to Wikipedia's NPOV style using the in-context learning capabilities of LLMs. Finally, the combined knowledge (existing and newly extracted content) in English is machine-translated into Hindi. We performed a comprehensive evaluation based on automated metrics and human assessments to demonstrate that the added content is informative, readable, and coherent. Our proposed pipeline is adaptable to any combination of HRL and LRL pairs. While the automated approach helps bridge information gaps in low-resource languages, it may risk overshadowing subtle cultural elements. To mitigate this, language-specific domain experts should perform thorough manual reviews before content integration. 

\section{Limitations}
Despite the promising results, our study has certain limitations. Our manual verification process, while crucial for ensuring content quality, is inherently subjective and may result in inconsistencies in evaluating relevance and accuracy. Furthermore, although the dataset of personal narratives is diverse, it may not fully represent all lesser-known biographies, which could limit the generalizability of our approach. Future research should aim to integrate more diverse sources and develop automated verification techniques to address these limitations.

\section{Ethical consideration}
The biographical writings used for data collection were sourced from publicly accessible digital libraries, adhering to copyright regulations and respecting intellectual property rights. All human annotators involved in the manual verification process participated voluntarily. No personally identifiable information was gathered from the annotators, ensuring their privacy and anonymity. Additionally, we took extensive measures to prevent the inclusion of sensitive or potentially harmful content in the enhanced Wikipedia articles.


\appendix
\label{sec:appendix}
\section{Prompt for generating neutral Wiki-style sentence}\label{ref:prompt_npov}
\begin{tcolorbox}[colback=red!5!white,colframe=red!75!black, title=Prompt for generating Wiki-style sentence]
\textbf{For each query message, remove framing bias and epistemological bias and do not add any extra content from your own knowledge.}

Framing bias: subjective words or one-sided words, revealing the author’s stance in a particular debate.

Epistemological bias: propositions that are either commonly agreed to be true or false and that are subtly presupposed, entailed, asserted or hedged in the text.

Here are some examples:.........

Provide only the Output as: <pad>output</pad>
\end{tcolorbox}

\section{Prompt for selecting the most relevant chunk to the given section content}\label{ref:prompt_chunk_select}
\begin{tcolorbox}[colback=red!5!white,colframe=red!75!black, title=Prompt for selecting most relevant chunk]
\textbf{For each query text, find out whether the given piece of text is relevant or not.}

start(*)

.......

end(*)

Evaluate whether the chunk between start(*) and end(*) is relevant to the given section content. A quality chunk should meet the following criteria: a) It should provide relevant information as compared with the content, b) it should be well-written.

Provide the output in the following format: 

--Yes/No

-- Confidence score: <score>

\end{tcolorbox}

\section{Groups in informativeness: FA category}
Given the high standard deviation observed in the informativeness metric for $c_{old}$, it is worthwhile to explore whether the improvements in content $c_{new}$ compared to $c_{old}$ are uniform across all the articles. We categorize the informativeness scores for $c_{old}$ into three ranges based on their distribution and record the corresponding scores for the same sections in $c_{new}$. Table~\ref{tab:auto_eval_FA_ranges} displays the informativeness scores for both $c_{old}$ and $c_{new}$ across these three groups. It is clear that the informativeness has improved in $c_{new}$ compared to $c_{old}$ in each group, mirroring the results shown in Table~\ref{tab:auto_eval_FA_ranges}.

\begin{table*}
    \centering
    \begin{tabular}{|c|c|c|c|}
    \hline
        Type & Group1 (0--50) & Group2 (50--100) & Group3 (100 and more) \\
        \hline
        $c_{old}$ & 18.6 (13.78) & 71.00 (13.92) & 177.18 (71.38) \\
        \hline
        $c_{new}$ & 61.79 (46.55) & 108.22 (50.07) & 235.18 (138.97) \\
        \hline
    \end{tabular}
    \caption{Automatic evaluation: mean and (standard deviation) of the metric informativeness divided into ranges of scores for the articles that belong to \textit{FA} quality class.}
    \label{tab:auto_eval_FA_ranges}
\end{table*}

\section{Human assessment for generated text}
The average assessment score for each evaluator is tabulated in Table~\ref{tab:human_eval}. Further, the evaluation is shown for FA and non-FA articles separately. 
\begin{table}[h]
    \scriptsize
    \centering
    \begin{tabular}{|p{12mm}|p{5mm}|p{5mm}|p{5mm}|p{5mm}|p{5mm}|p{5mm}|}
    \hline
    \textbf{Type} & \multicolumn{2}{c|}{\textbf{Informativeness}} & \multicolumn{2}{c|}{\textbf{Readability}} & \multicolumn{2}{c|}{\textbf{Coherence}} \\ \hline
    & FA & non-FA & FA & non-FA & FA & non-FA \\ \hline
    Evaluator 1 & 2.9 & 2.77 & 2.8 & 2.5 & 2.5 & 2.43 \\ \hline
    Evaluator 2 & 2.2 & 2.3 & 2.1 & 2.5 & 1.8 & 2.17\\ \hline
    Evaluator 3 & 3 & 2.93 & 2.4 & 2.33 & 2.3 & 2.0\\ \hline 
    Evaluator 4 & 2.6 & 2.7 & 2.6 & 2.7 & 2.6 & 2.53\\ \hline 
    Evaluator 5 & 2.4 & 2.37 & 1.9 & 2.17 & 2.0 & 2.03 \\ \hline 
    Evaluator 6 & 3 & 3 & 2.7 & 2.67 & 2.8 & 2.8 \\ \hline
    Evaluator 7 & 2.6 & 2.67 & 2.7 & 2.6 & 2.4 & 2.3 \\ \hline 
   \end{tabular}
    \caption{Human evaluation on the generated machine-translated Hindi content based on three metrics -- informativeness, readability, coherence.}
    \label{tab:human_eval}
\end{table}

\section{Old and newly generated content: sample sections}
\begin{figure*}
    \centering
    \includegraphics[height=10cm,width=\textwidth]{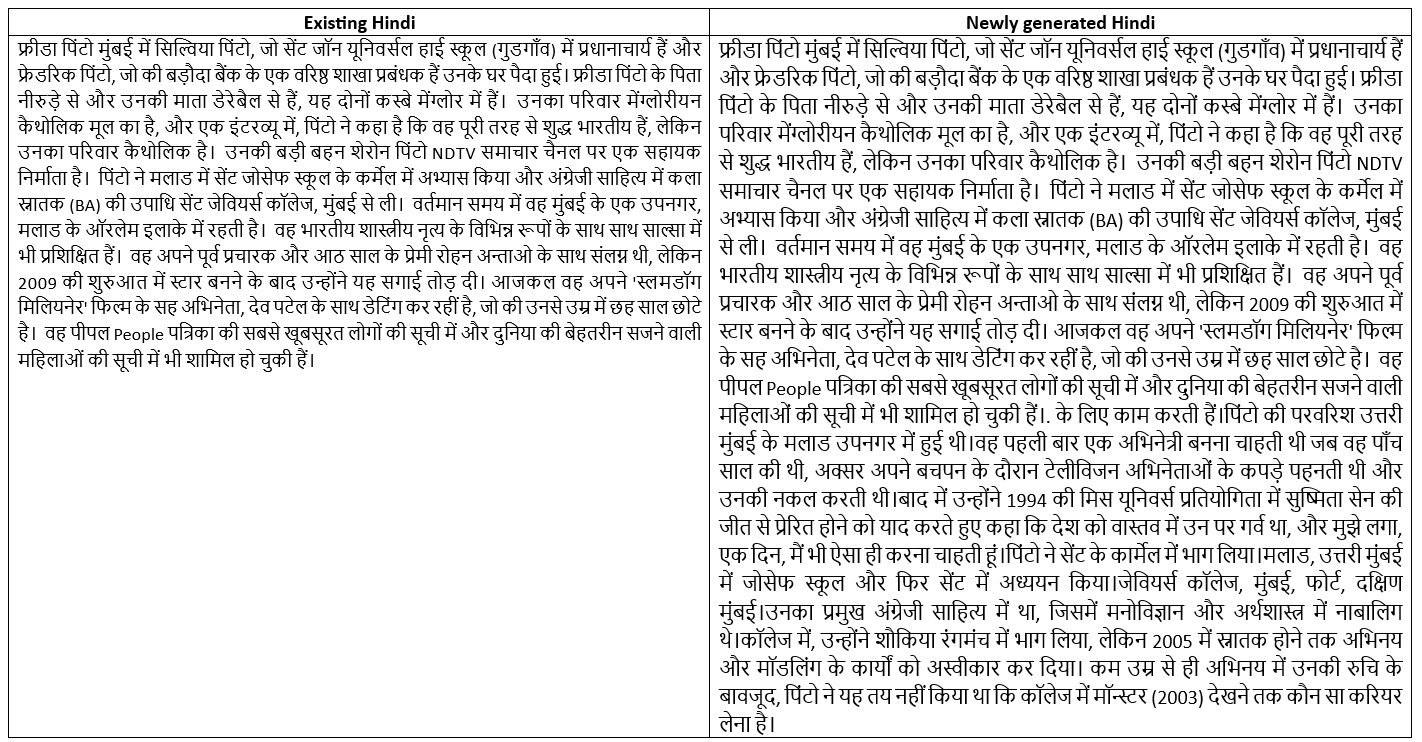}
    \caption{An example of existing and WikiTransfer generated new content -- a sample section that belongs to FA quality -- Hindi version.}
    \label{fig:sample_FA_hindi}
\end{figure*}

\begin{figure*}
    \centering
    \includegraphics[height=10cm,width=\textwidth]{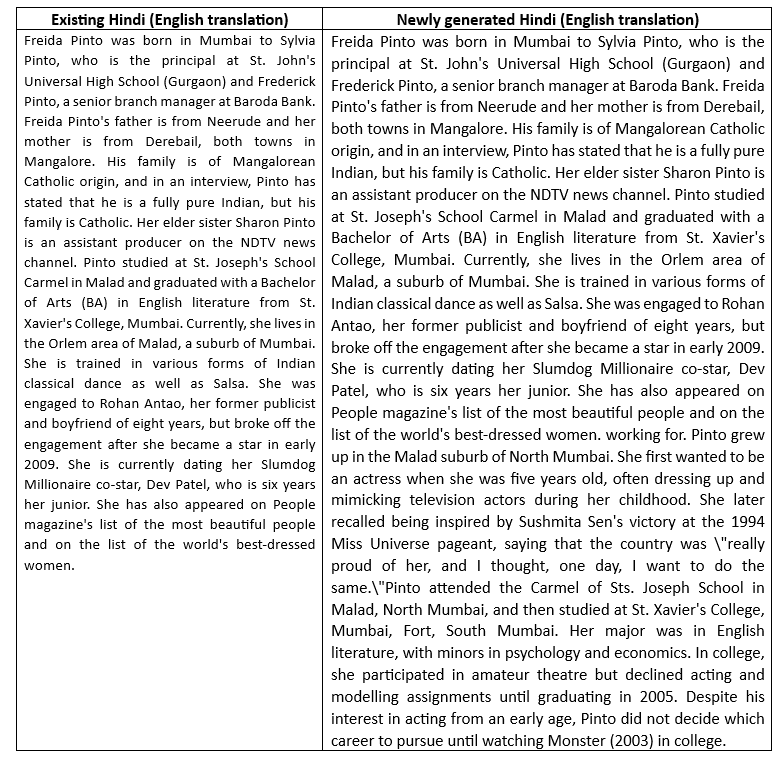}
    \caption{An example of existing and {WikiTransfer} generated new content -- a sample section that belongs to FA quality -- English version of Figure\ref{fig:sample_FA_hindi}.}
    \label{fig:sample_FA_english}
\end{figure*}

\begin{figure*}
    \centering
    \includegraphics[height=10cm,width=\textwidth]{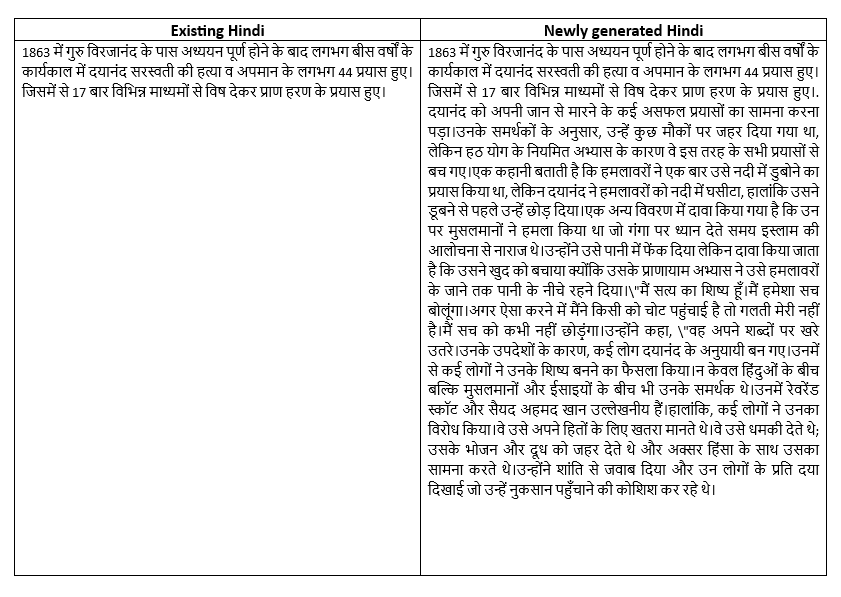}
    \caption{An example of existing and our framework generated new content-- a sample section that belongs to C quality -- Hindi version.}
    \label{fig:sample_C_hindi}
\end{figure*}

\begin{figure*}
    \centering
    \includegraphics[height=10cm,width=\textwidth]{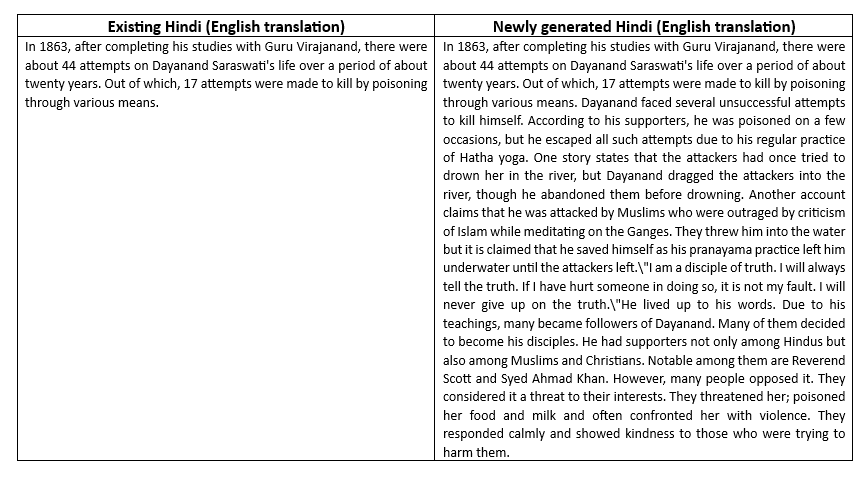}
    \caption{An example of existing and our framework generated new content-- a sample section that belongs to C quality -- English version of Figure~\ref{fig:sample_C_hindi}.}
    \label{fig:sample_C_english}
\end{figure*}

 The Hindi output for the FA article is generated using WikiTransfer, and both the Hindi content and its English translation are displayed in Figure~\ref{fig:sample_FA_hindi} and Figure~\ref{fig:sample_FA_english}. Similarly, for the C-class article, the new Hindi content is first pooled into text from external resources using the RAG method, followed by the NPOV correction and the WikiTransfer framework. The corresponding Hindi output and its English translation for this sample section are presented in Figure~\ref{fig:sample_C_hindi} and \ref{fig:sample_C_english}, respectively. 

\end{document}